\documentclass{IOS-Book-Article}

\usepackage{mathptmx}
\usepackage{soul}\setuldepth{article}
\usepackage{tabularx}
\usepackage{multirow}
\usepackage{graphicx}
\def\hb{\hbox to 11.5 cm{}}

\begin{document}

\pagestyle{headings}
\def\thepage{}
\begin{frontmatter}              

\title{Do LLMs Truly ``Understand'' When a Precedent Is Overruled?}

\markboth{}{September 2025\hb}


\author[A]{\fnms{Li} \snm{Zhang}\orcid{0000-0003-0375-1793}%
\thanks{Corresponding Author: Li Zhang, liz239@pitt.edu \\ The full dataset can be accessed at \url{https://github.com/lizhang-AIandLaw/Do-LLMs-Truly-Understand-When-a-Precedent-Is-Overruled}}},
\author[B]{\fnms{Jaromir} \snm{Savelka}\orcid{0000-0002-3674-5456}}, and
\author[A]{\fnms{Kevin} \snm{Ashley}\orcid{0000-0002-3674-5456}}

\runningauthor{L. Zhang et al.}
\address[A]{University of Pittsburgh}
\address[B]{Carnegie Mellon University}

\begin{abstract}
Large language models (LLMs) with extended context windows show promise for complex legal reasoning tasks, yet their ability to understand long legal documents remains insufficiently evaluated. Developing long-context benchmarks that capture realistic, high-stakes tasks remains a significant challenge in the field, as most existing evaluations rely on simplified synthetic tasks that fail to represent the complexity of real-world document understanding. Overruling relationships are foundational to common-law doctrine and commonly found in judicial opinions. They provide a focused and important testbed for long-document legal understanding that closely resembles what legal professionals actually do. We present an assessment of state-of-the-art LLMs on identifying overruling relationships from U.S. Supreme Court cases using a dataset of 236 case pairs. Our evaluation reveals three critical limitations: (1) \emph{era sensitivity} -- the models show degraded performance on historical cases compared to modern ones, revealing fundamental temporal bias in their training; (2) \emph{shallow reasoning} -- models rely on shallow logical heuristics rather than deep legal comprehension; and (3) \emph{context-dependent reasoning failures} -- models produce temporally impossible relationships in complex open-ended tasks despite maintaining basic temporal awareness in simple contexts. Our work contributes a benchmark that addresses the critical gap in realistic long-context evaluation, providing an environment that mirrors the complexity and stakes of actual legal reasoning tasks.

\end{abstract}

\end{frontmatter}
\markboth{September 2025\hb}{September 2025\hb}

\section{Introduction}
The recent expansion of the context windows of large language models (LLMs) has opened up possibilities for their application in domains that require understanding of long, complex documents \cite{liu2025comprehensive}. Transformer-based models \cite{vaswani2017attention} have evolved from the original 512-token limit to models capable of processing millions of tokens \cite{liu2024deepseek}, with key developments including sparse attention mechanisms \cite{han2023lm,tang2024quest} and memory-efficient attention algorithms \cite{liu2024deepseek}. Alternative approaches have been proposed to extend long-context modeling without changing the model architecture, such as retrieval-augmented generation (RAG) \cite{shi2023replug,gao2023retrieval} and the use of external memory \cite{wang2023augmenting}.

However, developing effective benchmarks for evaluating long-context understanding remains a significant challenge in the field. Evaluations relying on synthetic tasks, simplified question-answering formats, or artificially constructed scenarios fail to capture the complexity and stakes of real-world document understanding tasks. The legal field, with its reliance on lengthy case files and intricate statutes \cite{ruhl2007law}, presents an ideal domain. This paper presents an investigation into the capabilities of long-context LLMs in a particularly consequential legal task: the identification of overruling relationships. In U.S. law, overruling occurs when a later court declares that a prior decision should no longer be followed as binding precedent; we adopt this as the operational definition for this study.


We introduce a new dataset of 236 U.S. Supreme Court case pairs from 1789 to 2025, each consisting of an overruling case and the case(s) it overruled. Using this dataset, we evaluate a selection of long-context LLMs on three distinct tasks designed to probe the limits of their comprehension. We provide the models with the full texts of the overruling case which explicitly states that it overruled the specific case(s), forcing them to navigate a long context to identify the overruling relationship between them.

This task is challenging even when judicial opinions explicitly use terms like ``overruled.'' The underlying reasoning process is distributed across pages of text, requiring readers to integrate complex legal arguments across sections into a coherent logical chain. Judicial opinions are filled with specialized legal terminology and subtle nuances that require deep contextual understanding \cite{ashley2017artificial}. Our work addresses the critical gap in long-context evaluation by providing a benchmark that captures the realistic complexity of legal reasoning while maintaining the scientific rigor needed for meaningful model assessment.


\section{Related Work}
\label{sec:related}

\subsection{Long-Context Language Model Evaluation}

The evaluation of long-context models has led to the development of specialized benchmarks. LongBench \cite{bai2024longbench} introduced a comprehensive evaluation suite covering multiple languages and domains, while LongEval \cite{krishna2023longeval} and GSM-Infinite \cite{zhou2025gsm} focused specifically on the degradation of performance as context length increases. The SCROLLS benchmark \cite{shaham2022scrolls} provided domain-specific evaluation across various long-document tasks. LongDocURL \cite{deng2024longdocurl} and MMLONGBENCH-DOC \cite{ma2024mmlongbench} built multimodal long document benchmarks integrating understanding and reasoning. These benchmarks have revealed that performance often degrades significantly when information is embedded deep within long contexts, particularly for tasks requiring complex reasoning rather than simple information retrieval.

Our work extends this line of research by introducing a novel legal reasoning task that specifically tests models' ability to identify overruling relationships across lengthy judicial opinions. Unlike previous benchmarks that focus on general document understanding, our task requires temporal reasoning, hierarchical relationship identification, and the integration of complex legal arguments distributed across extended text.

\subsection{LLMs Application and Evaluation in Legal Domain}

The application of LLMs in the legal domain has seen significant growth, driven by the promise of automating complex legal tasks \cite{savelka2023unreasonable,gray2025generating} and improving access to legal information \cite{sovrano2024improve}. Several legal evaluation benchmarks have emerged to assess LLM performance in legal tasks. The LexGLUE benchmark \cite{chalkidis2021lexglue} provides a unified evaluation framework covering multiple legal NLP tasks including case outcome prediction, legal text classification, and legal question answering. The CUAD dataset \cite{hendrycks2021cuad} focuses specifically on contract understanding and analysis, testing models' ability to extract and reason about complex contractual terms. Other benchmarks \cite{guha2023legalbench,zhang2025measuring,magesh2025hallucination,pipitone2024legalbench} offer evaluation methods covering reasoning, information extraction, and document analysis tasks across various legal domains. These benchmarks have demonstrated that while LLMs show promising results on some legal tasks, they struggle with complex reasoning that requires deep understanding of legal principles.

Our work contributes to this body of research by introducing novel tasks that specifically test LLMs' ability to identify overruling relationships. Previous evaluations also mainly focus on multiple choice QA and classification tasks while our study tests the models' ability to navigate real-world full-context judicial opinions by designing three tasks with a mixture of open-ended and closed-ended questions.

\section{Dataset}
\label{sec:dataset}

To evaluate the performance of LLMs on the tasks of identifying overruled precedents, we created a new dataset of 236 U.S. Supreme Court case pairs in the period from 1789 to 2025. Each pair consists of a case that was explicitly overruled and the subsequent decision(s) that overruled it.

The foundation of our dataset is a resource maintained by the Constitution Annotated \cite{cong_rsch_serv_constitution_2025} consisting of 236 U.S. Supreme Court case pairs. The compilation process involved a comprehensive search of the Lexis database for all Supreme Court decisions containing the word ``overrule'' in the headnotes, syllabus, or opinion. A decision is included only when a majority of the Supreme Court has explicitly stated that it is being overruled. This strict criterion avoids the ambiguity inherent in legal commentary and ensures that each case in our dataset has been definitively identified as overruled by the Court itself. Conversely, cases that are merely distinguished, limited, or discredited without an explicit overruling are excluded. This conservative approach ensures the dataset's reliability. The results were then manually reviewed to confirm the Court's intent and to ensure that each case met the strict inclusion criteria.

For each overruling case of the 236 case pairs, we retrieved the full opinion texts from CourtListener \cite{freelawproject2020recap}, including majority and lead opinions as well as combined, concurring, and dissenting opinions. We then cleaned the raw downloads to retain only the textual content and conducted a manual review to verify content quality and completeness. The resulting curated corpus constitutes the dataset used in this study.

\begin{figure}[hbt]
\centering
\includegraphics[width=0.8\textwidth]{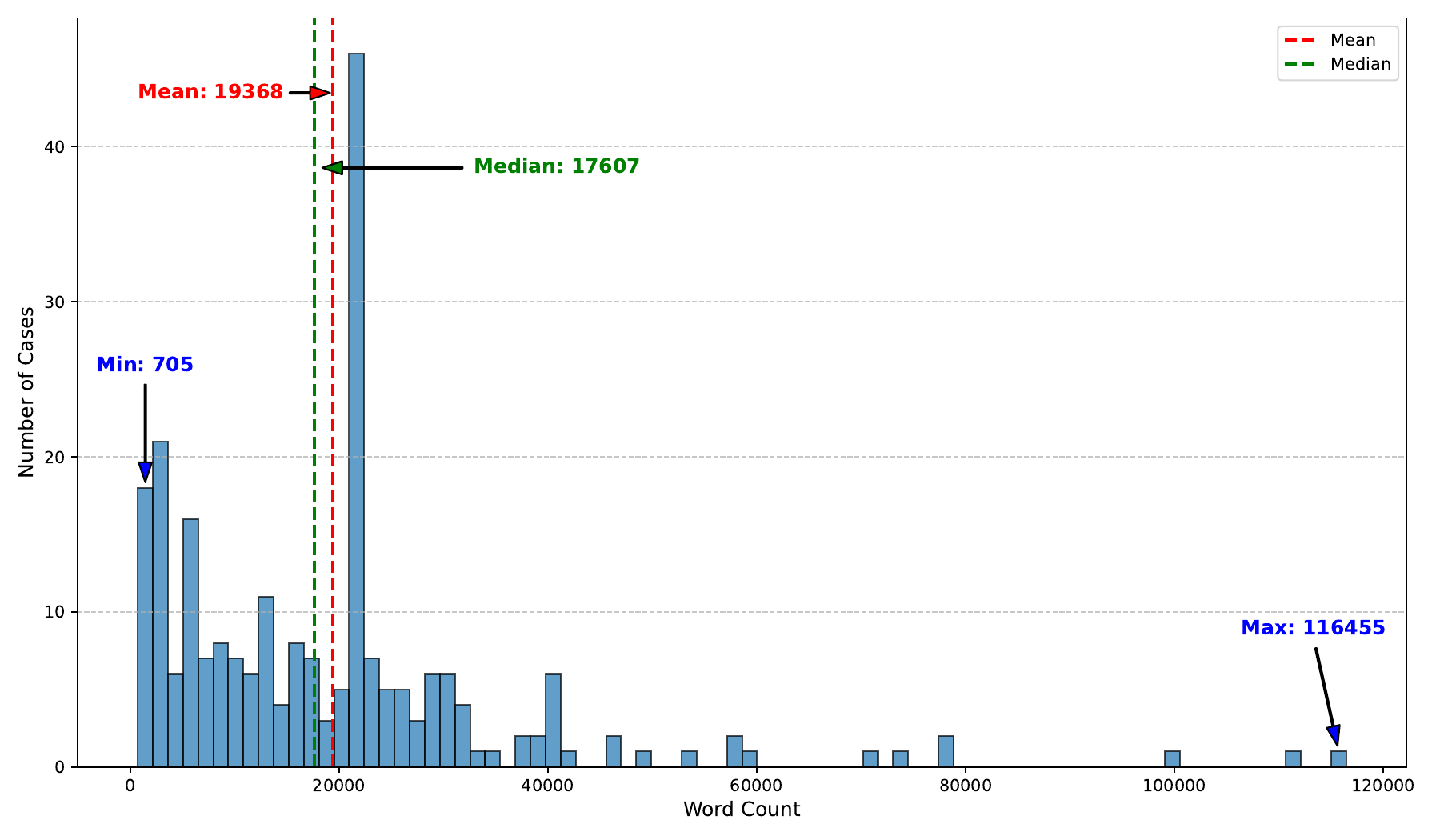}
\caption{Word Count Distribution of 236 Overruling Cases in Our Dataset.}
\label{fig:word_count}
\end{figure}

Figure~\ref{fig:word_count} illustrates the distribution of word counts across our dataset. The 236 overruling cases range from the shortest case with 705 words to the longest with 116,455 words. The median case length of 17,607 words and mean case length of 19,368 suggest that most cases in our dataset require efforts to process and understand, making them ideal for testing the limits of LLM long-context comprehension.

\section{Experiment Design}
\label{sec:expdesign}

To probe the capabilities of LLMs' long-context understanding in the legal domain, we designed three tasks centered on identifying overruling relationships.

\subsection{Task 1: Open-Ended Identification}
The first task poses an open-ended question to LLMs: ``Which case(s) was overruled by [the overruling case, e.g., ``\emph{Hohn v. United States}, 524 U.S. 236 (1998)'']? Here is the related context [case text of the overruling case].'' In this prompt, we use the overruling case as the context and ask the LLM to identify the case(s) that were overruled by the overruling case.

\paragraph{Rationale} This task simulates a realistic legal research query. To evaluate the open-ended responses, we employed \texttt{GPT-4o} as an automated judge, using a few-shot prompt to ensure consistent and accurate assessment against the ground truth. The core prompt was designed as ``Do [the model's response, e.g., ``\emph{House v. Mayo}''] and the ground truth [the overruled case(s), e.g., ``\emph{House v. Mayo}, 324 U.S. 42 (1945)''] refer to the same case?''

The development of the few-shot examples involved two rounds of iterative testing. First, we compared GPT-4o's zero-shot judgment results against our ground truth annotations and analyzed the discrepancies to identify patterns in misclassification. Based on these findings, we refined the prompt by incorporating carefully selected examples that addressed the common failure modes. In the second round, we conducted a validation experiment where 10\% of the responses were randomly selected for human expert review, demonstrating that GPT-4o's accuracy as a judge under the few-shot prompting approach was 98\% consistent with human expert assessments.

\subsection{Task 2: Closed-Ended Verification}
The second task is a verification query: ``Did [the overruling case, e.g., ``\emph{Hohn v. United States}, 524 U.S. 236 (1998)''] overrule [the overruled case, e.g., `` \emph{House v. Mayo}, 324 U.S. 42 (1945)'']? Here is the related context [case text of the overruling case].'' With this prompt, the model must answer ``true,'' ``false,'' or ``unknown.'' The correct answer is always ``true.''

\paragraph{Rationale} This task reduces the problem from open-ended QA to closed-ended verification. It allows us to isolate the model's ability to confirm or deny a specific legal relationship when presented with all necessary context, testing its comprehension and reasoning over long distances within the text.

\subsection{Task 3: Reversed Closed-Ended Verification}
The third task presents a logically flawed question: ``Did [the overruled case, e.g., `` \emph{House v. Mayo}, 324 U.S. 42 (1945)''] overrule [the overruling case, e.g., ``\emph{Hohn v. United States}, 524 U.S. 236 (1998)'']? Here is the related context [case text of the overruling case].'' With this prompt, the model must answer ``true,'' ``false,'' or ``unknown.'' The year of each case is provided in the context. The correct answer is always ``false,'' as a precedent can only be overruled by a later decision.

\paragraph{Rationale} This task serves as a control to test whether failures in the other tasks stem from a fundamental misunderstanding of temporal logic or from difficulties in understanding the long context. A model that understands chronology should consistently answer ``false''. 

\subsection{Models}
We evaluated five state-of-the-art LLMs known for their strong performance on long-context benchmarks. At the time of our experiments, the Gemini-Pro \cite{google_modelcards}, Gemini-Flash, Qwen3 \cite{yang2025qwen3} represented the top 3 of the LongBench v2 leaderboard \cite{bai2024longbench}, and were selected to ensure our evaluation reflected the current state of the art. We also included GPT-5 \cite{openai_gpt5_system_card} and Gemini-Flash-Lite to broaden our assessment. The models and their context window length (in tokens) are:
\begin{itemize}
    \item \texttt{Qwen3-235B-A22B-Thinking-2507} (Qwen3) - Context Length: 262,144
    \item \texttt{Gemini-2.5-flash-lite-preview-06-17} (Gemini-Flash-Lite) - Context Length: 1,000,000
    \item \texttt{Gemini-2.5-flash-preview-05-20} (Gemini-Flash) - Context Length: 1,000,000
    \item \texttt{Gemini-2.5-pro-preview-06-05} (Gemini-Pro) - Context Length: 1,000,000
    \item \texttt{GPT-5-2025-08-07} (GPT-5) - Context Length: 128,000
\end{itemize}

\subsection{Experimental Setup}
All experiments were conducted with a \texttt{temperature} of 0.1, a \texttt{top\_p} of 0.5, and a structured json output format, to encourage deterministic responses. For GPT-5 model, we use the default parameters as set by OpenAI, as the relevant parameter controls are not publicly exposed.

\section{Empirical Results}
\label{sec:results}

The performance of the selected LLMs is summarized in Table~\ref{tab:results}.

\begin{table}[hbt]
\caption{Model Performance Across Tasks (Accuracy in \%, higher is better; $\pm$ denotes 95\% confidence intervals, $n=236$).}
\label{tab:results}
\begin{tabular}{@{}l c@{\hspace{1.5cm}}c@{\hspace{1.5cm}}c@{}}
\hline
\textbf{Model} & \textbf{Task 1} & \textbf{Task 2} & \textbf{Task 3} \\
\hline
Qwen3 & 2.12\% $\pm$ 1.84\% & 1.27\% $\pm$ 1.42\% & 97.46\% $\pm$ 2.02\% \\
Gemini-Flash-Lite & 0.85\% $\pm$ 1.01\% & 11.86\% $\pm$ 4.08\% & 91.95\% $\pm$ 3.52\% \\
Gemini-Flash & 33.47\% $\pm$ 6.03\% & 6.36\% $\pm$ 3.12\% & 62.71\% $\pm$ 6.12\% \\
Gemini-Pro & \textbf{73.31\% $\pm$ 5.65\%} & 42.37\% $\pm$ 6.25\% & \textbf{100\% $\pm$ 0\%} \\
GPT-5 & 71.19\% $\pm$ 5.78\% & \textbf{75.42\% $\pm$ 5.47\%} & \textbf{100\% $\pm$ 0\%} \\

\hline
\end{tabular}
\end{table}

\subsection{Task 1: Open-Ended Identification}

In the open-ended identification task, the models struggled to correctly identify the overruled case. The best-performing model, Gemini-Pro, achieved an accuracy of 73.31\% (173 out of 236), closely followed by GPT-5 with 71.19\% (168 out of 236). The other models performed worse. This suggests that even state-of-the-art models have difficulty with understanding the overruling relationship in a long legal context.

There are three patterns in the errors. First, when an overruling case overturns multiple precedents, models often extract only one of them, yielding a partially correct answer. Second, models sometimes confuse cases that are merely mentioned in the text with the case actually overruled---we refer to this as \emph{confusion} errors, where models incorrectly identify a case that appears in the text but is not the overruled precedent. Third, models hallucinate a case that is not mentioned in the opinion at all---we define \emph{hallucination} as the generation of case names or citations that do not exist in the provided context; for example, in Gemini-Flash-Lite's response to \emph{Garland v. Washington}, the model produced \emph{Miles v. Graham}, which is neither cited in the text nor overruled by \emph{Garland}. In the calculation of the accuracy, we only count the cases where the model correctly identified all overruled cases. Table~\ref{tab:task1_examples} provides illustrative examples of these patterns. 

\begin{table}[hbt]
\caption{Examples of Model Responses and Error Patterns for Task 1}
\label{tab:task1_examples}
\begin{tabularx}{\textwidth}{X X X X@{}}
\hline
\textbf{Overruling Case} & \textbf{Overruled Case} & \textbf{Model Response} & \textbf{Type} \\
\hline
\emph{Hohn v. United States}, 524 U.S. 236 (1998) & \emph{House v. Mayo}, 324 U.S. 42 (1945) & \emph{House v. Mayo}, 324 U.S. 42 (1945) & Correct \\
\emph{Payne v. Tennessee}, 501 U.S. 808 (1991) & \emph{South Carolina v. Gathers}, 490 U.S. 805 (1989) \& \emph{Booth v. Maryland}, 482 U.S. 496 (1987) & \emph{South Carolina v. Gathers}, 490 U.S. 805 (1989) & Missing Case \\
\emph{Garland v. Washington}, 232 U.S. 642 (1914)  &  \emph{Crain v. United States}, 162 U.S. 625 (1896) &  \emph{Rogers v. Peck}, 199 U. S. 425, 435 (1905) & Confusion \\
\emph{Garland v. Washington}, 232 U.S. 642 (1914)  &  \emph{Crain v. United States}, 162 U.S. 625 (1896) &  \emph{Miles v. Graham}, 268 U.S. 501 (1925) & Hallucination \\
\hline
\end{tabularx}
\end{table}

\subsection{Task 2 \& 3: Closed-Ended Verification}

The closed-ended verification of Task 2 yielded more varied results. GPT-5 emerged as the top performer in this task, achieving an accuracy of 75.42\% (178 out of 236), significantly outperforming Gemini-Pro which achieved 42.37\% (100 out of 236). GPT-5 also showed a much lower rate of abstention, answering ``unknown'' in only 3 cases compared to Gemini-Pro's 131 abstentions. However, this low abstention rate comes with a significant cost: GPT-5 incorrectly labeled 55 cases as ``False,'' representing a substantial number of false negatives.

While GPT-5's low abstention rate might suggest higher confidence in its reasoning capabilities, it could also be interpreted as a tendency toward overconfident incorrectness. This trade-off highlights the need for more sophisticated uncertainty quantification mechanisms that can distinguish between genuine epistemic uncertainty and mere comprehension gaps. The other models showed a significant drop in performance, with Gemini-Flash answering ``unknown'' in the majority of cases. This suggests that even when the task is simplified to verification, most models struggle to confidently identify the correct legal relationship.

The reversed closed-ended verification of Task 3 showed a marked improvement in performance across all models. The accuracy was higher than in Task 2. GPT-5 and Gemini-Pro achieved high accuracy (100\%), correctly identifying the logical impossibility in all cases. Qwen3 also performed well with 97.5\% accuracy.

\subsection{Discussion}

\subsubsection{Era Sensitivity}

To probe temporal robustness in the open-ended identification task (Task 1), we stratified cases by overruling case decision year (the earliest overruling case happened in 1810) and measured models' accuracy across the fixed historical intervals that reflect changes in case density and legal language evolution. There are 17 case pairs from 1810--1881, 65 case pairs from 1882--1953, and 154 case pairs from 1954--2025. This quantitative view in Table~\ref{tab:error-by-fixed-intervals} aligns with our qualitative observations and clarifies where failures concentrate.

\begin{table}[hbt]
\caption{Model Performance by Historical Intervals in Task 1 (Accuracy in \%, higher is better).}
\label{tab:error-by-fixed-intervals}
\centering
\begin{tabular}{@{}l c@{\hspace{1.5cm}}c@{\hspace{1.5cm}}c@{}}
\hline
\textbf{Model} & \textbf{1810--1881} & \textbf{1882--1953} & \textbf{1954--2025} \\
\hline
Qwen3 & 0.00\% & 1.54\% & 2.60\% \\
Gemini-Flash-Lite & 0.00\% & 0.00\% & 1.30\% \\
Gemini-Flash & 17.65\% & 24.62\% & 38.96\% \\
Gemini-Pro & \textbf{64.71\%} & \textbf{72.31\%} & 74.68\% \\
GPT-5 & 35.29\% & 53.85\% & \textbf{82.47\%} \\
\hline
\end{tabular}
\end{table}

The temporal stratification reveals era sensitivity across all models, with performance degrading as we move backward in time. In the earliest period (1810--1881), the performance gap is stark: Qwen3 and Gemini-Flash-Lite achieve 0\% accuracy, completely failing to identify any overruling relationships in this historical interval. The best-performing model in this period, Gemini-Pro, manages only 64.71\% accuracy, while GPT-5 struggles at 35.29\%. This performance drop in the earliest cases suggests that models trained primarily on modern legal texts lack the linguistic and conceptual frameworks needed to parse 19th-century judicial opinions, which often employ archaic legal terminology, different citation conventions, and distinct argumentative structures \cite{mellinkoff2004language}.

The middle period (1882--1953) shows a moderate improvement. Here, Gemini-Pro maintains its strong performance at 72.31\% accuracy, demonstrating remarkable temporal consistency. GPT-5 shows significant improvement to 53.85\%, while Gemini-Flash reaches 24.62\%. However, the weaker models continue to struggle: Qwen3 achieves only 1.54\% accuracy, and Gemini-Flash-Lite reaches 0.00\%. This intermediate period represents a transitional phase where legal language begins to modernize but still retains many historical characteristics that challenge contemporary models. The performance improvements suggest that models can adapt to some degree of historical variation, but their capabilities remain fundamentally limited when dealing with pre-modern legal discourse.

The modern period (1954--2025) represents the models' comfort zone, with significantly improved performance across the board. GPT-5 achieves its highest accuracy at 82.47\%, significantly outperforming Gemini-Pro's 74.68\%. Gemini-Flash shows improvement to 38.96\% while remaining limited. This improvement in the modern era reflects several factors: the models' training data heavily emphasizes contemporary legal texts and the legal reasoning patterns align more closely with what the models have learned during training \cite{gao2020pile}. The fact that even the best-performing models achieve only around 80\% accuracy underscores the fundamental challenge of legal reasoning tasks, while the precipitous drop in performance for earlier periods quantifies the temporal bias that underlies the open-ended reasoning failures we observed in our experiments.

\subsubsection{Hesitation in Verification vs. Confidence in Refutation}

\begin{figure}[hbt]
\centering
\includegraphics[width=0.8\textwidth]{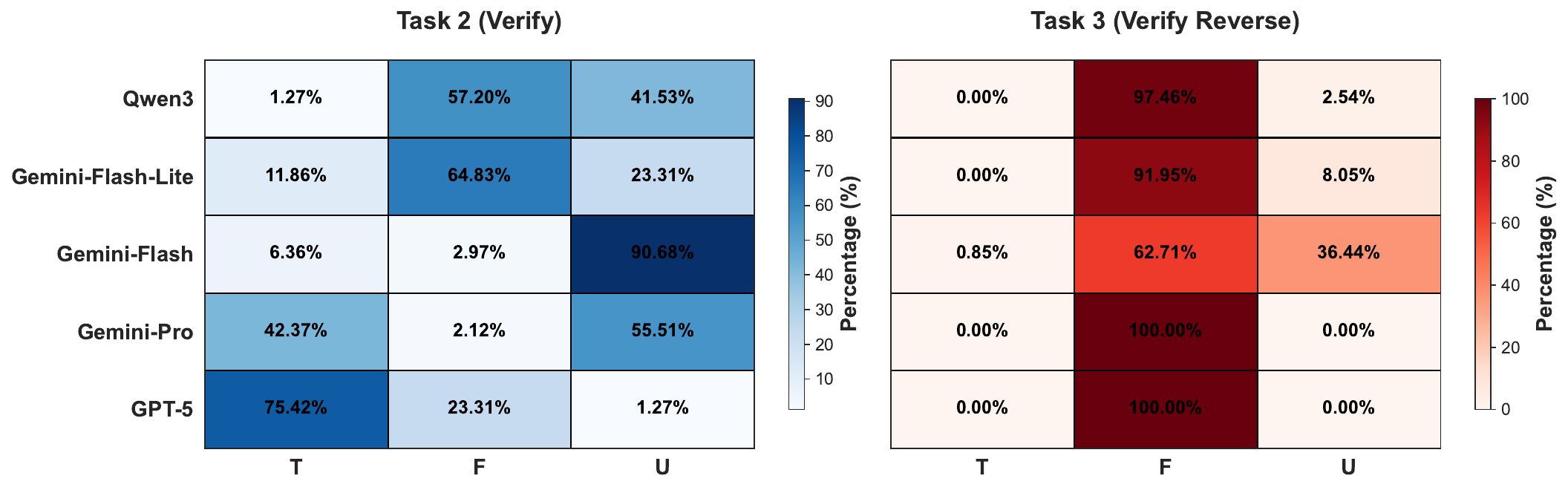}
\caption{Confusion Matrix for Verification Tasks. T, F, and U stand for True, False, and Unknown respectively.}
\label{fig:confusion_matrix}
\end{figure}

The high number of ``unknown'' responses in Task 2 is revealing, as shown in Figure~\ref{fig:confusion_matrix}. It suggests that most models are not simply guessing, but are actively assessing their own uncertainty. When the models are unable to reason and understand the case relationship embedded in the text, they are hesitant to make a definitive judgment. This is a desirable trait in a legal AI system, as it is preferable for a model to admit its own ignorance rather than to provide a confident but incorrect answer. However, the high rate of abstention in this task suggests that most models' threshold for certainty is set too high, preventing them from making correct judgments even when the evidence is strong. GPT-5's low abstention rate (1.27\%, 3 out of 236) stands in stark contrast to this pattern, suggesting a different approach to uncertainty assessment.

A contrast emerges when comparing the results of Task 2 and Task 3. In Task 2, when asked to verify a correct overruling relationship, the models frequently defaulted to ``unknown,'' even when the answer was clearly stated in the text. In contrast, when presented with the temporally impossible proposition in Task 3, the same models answered with high confidence and accuracy, with GPT-5 and Gemini-Pro achieving 100\% accuracy and Qwen3 achieving 97.46\%.

This pattern of behavior reveals a fundamental limitation in the models' approach to legal reasoning. The contrast between Task 2 and Task 3 performance demonstrates that models rely on shallow logical heuristics rather than deep legal comprehension. In Task 3, models can successfully apply simple temporal logic—recognizing that an earlier case cannot overrule a later one—without actually understanding the complex legal reasoning embedded in the judicial opinions. This high performance is achieved through basic chronological awareness rather than genuine comprehension of legal arguments.

\subsubsection{Temporal Reasoning}

Our analysis also reveals an interesting pattern in temporal reasoning across different task contexts. In the open-ended identification task (Task 1), we observed what we term ``context-dependent temporal reasoning failures,'' as illustrated in Table~\ref{tab:temporal_reasoning_failures}. These failures show that models can create temporally impossible relationships, such as suggesting that a 1914 case overruled a 1925 case, even when the case names include the years. Such responses indicate a breakdown in temporal causality, where a case from an earlier year is incorrectly identified as overruling a case from a later year.

\begin{table}[hbt]
    \caption{Examples of Context-Dependent Temporal Reasoning Failures from Task 1. LLMs generated logically impossible scenarios where a subsequent case was overruled by a preceding one.}
    \label{tab:temporal_reasoning_failures}
    \begin{tabularx}{\textwidth}{@{}Xl@{}}
    \hline
    \textbf{Case Relationship / Query} & \textbf{Year} \\
    \hline
    \emph{Query Case (Overruling Case):} & \\
    \quad Garland v. Washington & 1914 \\
    \emph{Model's Response (Overruled Case):} & \\
    \quad Miles v. Graham & 1925 \\
    \hline
    \end{tabularx}
\end{table}

However, when we examine the results from Task 3 (the reversed verification task), a different picture emerges. As shown in Table~\ref{tab:results}, models demonstrate remarkably high accuracy in identifying temporally impossible propositions. GPT-5 and Gemini-Pro achieved 100\% accuracy, while Qwen3 achieved 97.46\% accuracy in correctly rejecting statements like "Did \emph{Booth v. Maryland} (1987) overrule \emph{Payne v. Tennessee} (1991)?"—a clear temporal impossibility.

This contrast suggests that the models are not fundamentally lacking in temporal reasoning capabilities or awareness of legal case chronology. Rather, the issue lies in the complexity of the reasoning required in different contexts. In Task 3, the models are presented with a straightforward logical contradiction that they can easily identify using basic temporal logic. In contrast, Task 1 requires more sophisticated reasoning—the models must understand complex legal relationships, extract relevant information from dense legal texts, and apply temporal logic in an open-ended context.

\section{Limitations and Future Work}
\label{sec:limitations}

Our study has several limitations. First, our dataset focuses exclusively on U.S. Supreme Court cases, which may not generalize to other jurisdictions or legal systems with different precedent structures. Second, we only examined cases with explicit overruling language, excluding more nuanced legal developments like distinguishing, limiting, or discrediting precedents. Third, our evaluation relies on automated assessment using GPT-4o, which, while validated against human experts, may introduce its own biases. Finally, we tested a limited set of models at a specific point in time, and rapid advances in LLM technology may render some findings obsolete.

Future work should address these limitations by expanding to international legal systems, incorporating more subtle legal relationships, and developing more sophisticated evaluation metrics. Additionally, research should explore architectural improvements for long-context understanding, such as legal domain-specific pre-training, better uncertainty quantification, and the potential of RAG systems \cite{li2024retrieval,qi2024long} to enhance LLMs' long-context understanding capabilities. Exploring agent-based approaches \cite{zhang2024chain,zhang2025mitigating} for enhancing LLMs' long-text capabilities could also lead to more consistent and reliable legal AI systems. The development of legal reasoning benchmarks that test deeper comprehension would further advance the field, along with more detailed quantitative analysis of error patterns across different models to strengthen qualitative insights into specific failure modes.


\section{Conclusion}
\label{sec:conclusion}

This paper presents an evaluation of state-of-the-art LLMs on the legal reasoning task of identifying overruling relationships from U.S. Supreme Court cases. Our investigation reveals three fundamental limitations that challenge the practical deployment of these models in legal contexts.

First, we demonstrate that models exhibit era sensitivity. The temporal stratification reveals era sensitivity across all models in our evaluation, with performance degrading as we move backward in time. This temporal bias reveals that extended context windows alone cannot overcome the fundamental challenge of understanding legal language across different historical periods, suggesting that models lack the linguistic and conceptual frameworks needed to parse judicial opinions from earlier eras that employ archaic legal terminology, different citation conventions, and distinct argumentative structures.

Second, our analysis reveals the models' reliance on shallow logical heuristics rather than deep legal comprehension. Models can successfully apply simple temporal logic—such as recognizing that an earlier case cannot overrule a later one—without actually understanding the complex legal reasoning embedded in the judicial opinions. The high performance on Task 3 is achieved through basic chronological awareness rather than genuine comprehension of legal arguments, highlighting a fundamental gap between surface-level logical reasoning and deep legal understanding.

Third, we identify context-dependent temporal reasoning failures that highlight the complexity of legal reasoning tasks. Models can correctly identify simple temporal contradictions but fail to maintain temporal consistency when processing complex, open-ended legal relationships. This finding indicates that the challenge lies not in basic logical reasoning, but in integrating temporal awareness with deep legal understanding under cognitive load.

The empirical results underscore a critical gap between current LLM capabilities and legal practice requirements. Even the best-performing model achieved only 73.31\% accuracy on the core identification task, falling short of the reliability standards needed for legal applications. This performance gap is particularly concerning given that our dataset focuses on cases with explicit overruling language, representing a straightforward scenario for legal relationship identification.

\bibliographystyle{plain}
\bibliography{reference}

\end{document}